\documentclass{article}
\usepackage{PRIMEarxiv}
\usepackage[utf8]{inputenc}
\usepackage[T1]{fontenc}
\usepackage{lmodern}
\usepackage{float}
\usepackage[table,dvipsnames]{xcolor}
\usepackage{amsmath,amssymb,amsfonts}
\usepackage{mathtools}
\usepackage{mathrsfs}
\usepackage{amsthm}
\usepackage[colorlinks=true]{hyperref}
\hypersetup{
  linkcolor=black,
  citecolor=black,
  urlcolor=[RGB]{5,115,185}
}
\usepackage{url}
\usepackage{graphicx}
\usepackage{bm}
\usepackage[ruled,vlined]{algorithm2e}
\usepackage{algpseudocode}
\usepackage{nicefrac}
\usepackage{microtype}
\usepackage{authblk}
\usepackage{fancyhdr}
\usepackage{booktabs}
\usepackage{enumitem}
\usepackage{multirow}
\usepackage{tabularx}
\usepackage{array}
\newcolumntype{L}{>{\raggedright\arraybackslash}X}
\newcolumntype{Y}{>{\centering\arraybackslash}X}

\usepackage{colortbl}
\usepackage{stmaryrd}
\usepackage{sidecap}
\usepackage{titlesec}
\usepackage{footmisc}
\usepackage[skip=6pt,justification=centering]{caption}
\usepackage{subcaption}
\usepackage{pgfplots}
\pgfplotsset{compat=1.16}
\usepackage{pgfplotstable}
\usepackage{pifont}
\usepackage{listings}

\usepackage[sorting=none, backend=biber]{biblatex} 
\definecolor{LightGray}{rgb}{0.8,0.8,0.8}
\definecolor{LightCyan}{rgb}{0.74,0.83,0.9}
\definecolor{DarkBlue}{rgb}{0,0.28,0.67}
\definecolor{blizzardblue}{rgb}{0.67, 0.9, 0.93}
\definecolor{inchworm}{rgb}{0.7, 0.93, 0.36}
\definecolor{coralred}{rgb}{1.0, 0.25, 0.25}
\definecolor{celadon}{rgb}{0.67, 0.88, 0.69}
 
\DisableLigatures{encoding = *, family = tt* }

\usepackage{xspace}
\newcommand{\sherpa}{\href{https://www.sherpa.ai/}{\textcolor{DarkBlue}{Sherpa.ai}}\xspace}
 
\definecolor{DarkColor}{gray}{0.75}
\definecolor{LightColor}{gray}{0.9}
\definecolor{LightGrey}{rgb}{0.98,0.98,0.98}
\definecolor{DarkGrey}{rgb}{0.83,0.83,0.83}
\definecolor{BaseColor}{rgb}{0.10,0.10,0.20}
\definecolor{TextColor}{RGB}{58,88,119}
\definecolor{LightTextColor}{RGB}{229,233,205}
\definecolor{DarkTextColor}{RGB}{25,29,1}
\definecolor{NeutralBg}{rgb}{0.92,0.92,0.92}
\definecolor{LightYellow}{RGB}{255,255,102}
\definecolor{DarkOrange}{RGB}{255,90,0}
\definecolor{Green}{RGB}{0,128,0}
\definecolor{White}{gray}{1}
 
\usepackage{etoolbox}
\makeatletter
\patchcmd{\@begintheorem}{\textit}{\textbf}{}{}
\makeatother

\numberwithin{equation}{section}
\numberwithin{subsection}{section}

\AtBeginEnvironment{tabular}{\small}
 
\usepackage{tikz}
\usepackage{quantikz}
\usepackage{dsfont}
\newcommand{\ind}{\mathds{1}}

\title{\textcolor{black}{Federating Quantum and Classical Computing: A Privacy-Preserving Hybrid Approach}}

\author{%
  {\LARGE \href{https://sherpa.ai/}{Sherpa.ai}}\\
  research@sherpa.ai
}

\fancypagestyle{firstpagestyle}{%
  \fancyhf{}%
  \fancyhead[R]{\includegraphics[scale=1.0]{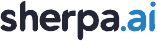}} 
  \fancyfoot[C]{\thepage}%
}

\begin{document}

\maketitle
\thispagestyle{firstpagestyle}

\begin{abstract}


Quantum machine learning (QML) is increasingly recognized as one of the most promising near-term applications of quantum computing, viewed as a next-frontier candidate beyond purely classical approaches. Hybrid quantum-classical models operationalize this potential by embedding a parameterized quantum circuit within a model where all other components remain classical---a design already applied to chemistry simulation, financial modeling, and image classification. However, their deployment in privacy-sensitive, multi-party settings is constrained by the need to avoid centralizing raw data and by the requirement that modern quantum circuits remain parameter-efficient to stay trainable at scale.

In this paper, we address these constraints by evaluating federated learning (FL) as a means of combining a hybrid quantum-classical active party with a classical passive party, using \sherpa's Blind Vertical FL (SBVFL) protocol to avoid centralizing raw data, while drastically reducing communication. We construct the split multiplicative periodic parity (SMPP) benchmark, following common QML design practice. On this task, our simulations show that SBVFL raises accuracy from 0.7227 to 0.8757 compared to local training, closely approaching non-private centralized accuracy, and that the hybrid quantum-classical model achieves this with substantially fewer trainable parameters than the classical neural networks and random forest alternatives. These results show that FL enables high-performing, privacy-preserving quantum-classical collaboration without centralizing raw data.

\end{abstract}

\begin{figure}[h]
  \centering
  \includegraphics[width=0.6\linewidth]{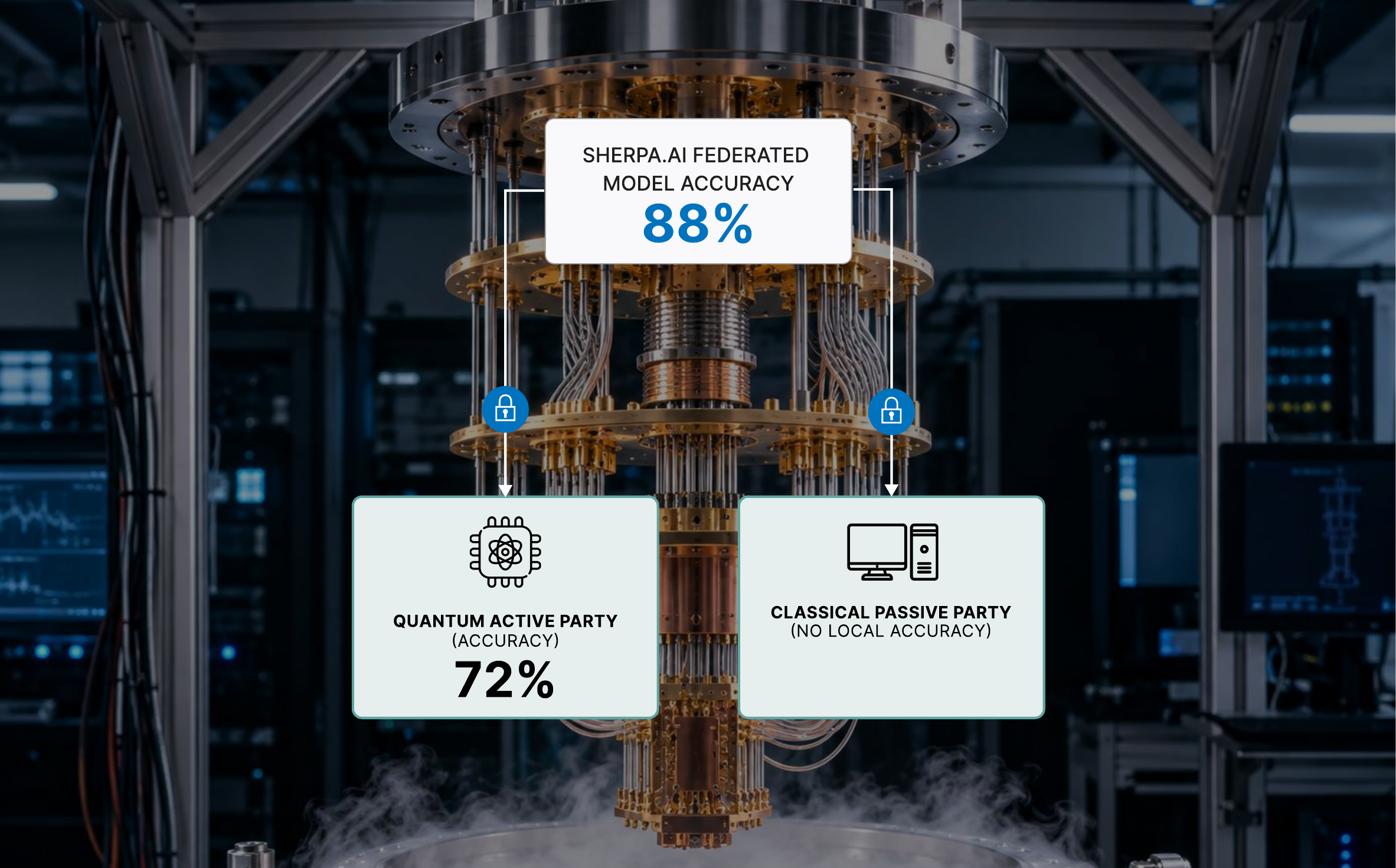}
  \caption{Overview of the FL setup showing the single-party and federated model performance.}
  \label{fig:compare_local_fl}
\end{figure}

\section{Introduction}
\label{sec:introduction}

Quantum machine learning (QML) has attracted growing attention as a leading near-term application of quantum computing, with the potential to extend beyond classical approaches. Quantum computing exploits superposition, entanglement, and interference to implement computational procedures that can differ fundamentally from those of classical processors. These resources can provide asymptotic advantages for specific problem classes, but they do not imply a general acceleration of machine-learning workloads. Shor's and Grover's algorithms illustrate this problem-specific character~\cite{nielsen2010quantum}, while QML uses quantum states and parameterized circuits to construct model classes with task-dependent inductive biases~\cite{biamonte2017quantum}.

Near-term quantum computing remains constrained by noise, limited coherence, gate infidelity, restricted device size, and the resulting limitations on circuit depth~\cite{preskill2018quantum}. Hybrid quantum-classical architectures address these constraints by assigning different stages of the learning pipeline to classical and quantum processors. In QML, a parameterized quantum circuit can be embedded as a layer within a conventional model, with classical computation handling preprocessing and optimization~\cite{schuld2021effect}. This hybrid pattern has already been applied across domains, including chemistry simulation~\cite{gircha2023hybrid}, financial modeling~\cite{raj2023quantum}, and image classification~\cite{senokosov2024quantum}.

A parameterized circuit with angle encoding does not constitute a generic universal approximation mechanism. Instead, it induces a structured family of trigonometric functions~\cite{schuld2021effect}. Its usefulness therefore depends on whether the target function contains structure that this family represents efficiently. This motivates a model-class comparison in which the target is deliberately matched to the functional form of the quantum readout.

Federated learning (FL) enables multiple parties to train a shared predictive system without centralizing their raw data~\cite{mcmahan2017communication}. Quantum FL (QFL) combines this distributed setting with quantum or hybrid models (see Figure~\ref{fig:compare_local_fl}). Here we consider the heterogeneous case in which only the active party employs a hybrid quantum-classical model and the passive party remains entirely classical. The data are vertically partitioned, so the parties hold complementary feature subsets for the same samples; this setting is formalized in Section~\ref{sec:problem-formulation}.

Current quantum hardware also imposes practical restrictions on data preprocessing, circuit size, and repeated execution. We therefore embed the quantum circuit in a classical model and evaluate it by exact state-vector simulation. This removes device noise and queueing effects from the present comparison while retaining an architecture that can in principle be executed on a quantum processor of the corresponding size.

A further constraint concerns trainability. With parameter-shift gradient estimation, each trainable circuit parameter contributes additional circuit evaluations at every optimization step. Parameter count therefore provides a direct proxy for the number of parameter-dependent evaluations required for training, and we use the number of trainable parameters required to reach a specified accuracy as the principal complexity measure. The proxy describes hardware execution with parameter-shift gradients; the simulations reported here instead differentiate automatically through an exact state vector, so the parameter count is not a measured cost in these experiments.

\subsection{Motivation}
\label{sec:motivation}

Vertical FL (VFL) is appropriate when parties hold different feature subsets for a common set of samples. In a hybrid QFL setting, this permits different parties to use different model classes, including a quantum-enabled model at one party and a classical model at the other. The methodological question is whether a privacy-preserving protocol can combine these complementary feature views without exposing raw features or repeatedly transmitting label-dependent training information. In standard VFL, the passive party may receive label-dependent gradients repeatedly. We instead employ \sherpa's Blind VFL (SBVFL) paradigm~\cite{acero2025sbvfl}, in which the passive party trains against synthetic targets and does not receive label-dependent gradients. This protocol is well suited to a heterogeneous setting because the parties may use different architectures without requiring any exchange or averaging of model weights.

This combination of a heterogeneous, blind vertical protocol with a compact quantum active party is not addressed by the existing literature. Vertical QFL work to date is homogeneous, requiring quantum resources at every party~\cite{ballester2025vertical,luo2026evidential}, so it does not isolate the parameter efficiency of a quantum model against classical alternatives under an explicit privacy mechanism. Conversely, hybrid QFL studies that do compare quantum and classical model classes typically operate in a horizontal, non-private setting~\cite{innan2024fedqnn,innan2024qfnnffd}, so they leave open whether any measured advantage survives the information restrictions that a blind vertical protocol imposes on the passive party. It is therefore unclear, from prior work alone, whether a parameter-efficiency advantage observed for a quantum active party under centralized or local training is preserved, reduced, or eliminated once that party must rely on the passive party's contribution through SBVFL rather than through direct access to the pooled features.

Accordingly, the objective of this work is twofold: to characterize a target structure for which a compact quantum model is parameter-efficient at the active party, and to determine whether that advantage persists under the privacy and information constraints imposed by blind vertical federation.

This paper studies the structural conditions under which a compact quantum layer can be parameter-efficient for the active party of a SBVFL system and quantifies the resulting federated accuracy.

\subsection{Contribution}
\label{sec:contribution}

The main contributions are as follows:

\begin{itemize}
    \item We evaluate a heterogeneous hybrid quantum-classical SBVFL scheme, with a hybrid quantum-classical active party and an entirely classical passive party. The reported federated configuration reaches 0.8757 accuracy with 12 trainable parameters, substantially improving on active-party local training (0.7227) and closely approaching the non-private centralized reference.
    
    \item We construct split multiplicative periodic parity (SMPP), a benchmark whose vertical partition makes each party's feature block informative but insufficient on its own, matched to a tensor-product quantum readout. On the pooled four-feature task, a 19-parameter hybrid model exceeds the tested classical networks and the 300-tree random forest by margins from $0.055$ to $0.144$.

    \item We ablate the passive party's privacy multiplier and show that increasing privacy does not reduce accuracy, indicating that SBVFL's accuracy gains hold even under stronger privacy guarantees.
\end{itemize}

The remainder of the paper is organized as follows. Section~\ref{sec:problem-formulation} formalizes the hybrid quantum-classical VFL setting and reviews related work. Section~\ref{sec:ml-solution} describes the centralized and FL protocols, including SBVFL. Section~\ref{sec:benchmark} introduces the SMPP benchmark and its vertical partition. Section~\ref{sec:models} specifies the model classes, Section~\ref{sec:experiments} gives the experimental protocol, and Section~\ref{sec:results} presents the results. Section~\ref{sec:discussion} discusses the scope and interpretation of the findings, followed by conclusions in Section~\ref{sec:conclusions}.

\section{Problem Formulation}
\label{sec:problem-formulation}
 
This section formalizes the hybrid quantum-classical VFL setting by specifying the computational roles of the two parties, reviewing the relevant literature, and defining the learning problem.
 
In the hybrid quantum-classical VFL setting considered here, the parties may employ different computational models. The active party uses a hybrid model containing a variational quantum circuit, while the passive party uses a classical model. This heterogeneity allows the two parties to retain architectures appropriate to their available computational resources while participating in the same federated prediction task.

The practical limitations of current quantum processors motivate the classical preprocessing layer and exact simulation in this study. The architecture is nevertheless defined so that the quantum layer can be transferred to a device with the required number of qubits and circuit depth.

\subsection{Related Work}
\label{sec:related-work}
 
We review work most closely related to QFL and, in particular, its vertical setting, emphasizing approaches that distribute feature subsets rather than samples across parties.

Recent surveys characterize QFL from complementary perspectives, including architectural taxonomies, NISQ constraints, privacy mechanisms, and security~\cite{zaman2025survey,mathur2025when,nguyen2025qflsurvey,sai2025qflarchitecture,ballester2025survey}. The present study considers a narrower setting: heterogeneous VFL in which only the active party uses a quantum model, the passive party remains classical, and label blinding is explicit.
 
Vertical QFL remains less explored than horizontal QFL. \textcite{ballester2025vertical} describe a fully quantum VFL architecture in which client and server models are variational quantum circuits operating on distributed feature subsets. \textcite{luo2026evidential} propose eviQVFL, in which quantum clients transmit local states through quantum teleportation and the server performs evidential fusion.

\begin{table}[h]
  \small
  \centering
  \renewcommand{\arraystretch}{1.3}
  \begin{tabularx}{\textwidth}{L|Y|Y|Y}
    \toprule
    \textbf{Dimension} & \textbf{This work (SBVFL)} & \textbf{\textcite{ballester2025vertical}} & \textbf{\textcite{luo2026evidential}} \\
    \midrule
    Party capability & Heterogeneous (1 hybrid quantum-classical, 1 classical) & Homogeneous (all quantum) & Homogeneous (all quantum) \\
    Inter-party channel & Classical (encoded outputs) & Classical (intermediate outputs) & Quantum (teleportation) \\
    Explicit privacy mechanism & Label blinding + privacy multiplier & Structural (private local models) & Evidential fusion \\
    Hardware requirement & Simulator; compatible with a corresponding real QPU & Simulator (NISQ-oriented design) & Requires quantum teleportation channel \\
    Validation domain & SMPP, a tabular benchmark with multiplicative periodic
structure & Handwritten-digit image classification & Image classification \\
    \bottomrule
  \end{tabularx}
  \caption{Comparison with the two closest vertical QFL schemes.}
  \label{tab:related-work-comparison}
\end{table}

Table~\ref{tab:related-work-comparison} positions the present study relative to these approaches. The proposed architecture is heterogeneous, with quantum resources required only at the active party and a fully classical passive party, and it uses explicit label blinding with a tunable privacy multiplier. The inter-party channel is therefore classical and does not require quantum communication. The quantum layer is also related in spirit to the federated quantum neural architectures of \textcite{innan2024fedqnn,innan2024qfnnffd}, although those studies address horizontal federation. We therefore characterize the present contribution conservatively as a heterogeneous, blind hybrid quantum-classical VFL scheme and do not claim priority for vertical QFL in general.

\subsection{Problem Definition}
\label{sec:prob-description}
 
We consider a dataset $\mathcal{D}=(X,Y)$ with $X\in\mathbb{R}^{M\times d}$ and binary labels $Y\in\{0,1\}^{M}$. The feature matrix is vertically partitioned as $X=[X_a,X_p]$, where $X_a$ and $X_p$ denote the feature blocks held by the active and passive parties, respectively. Joint training uses only records matched by the Private Set Intersection (PSI) protocol~\cite{sherpapsi2023}; the common sample count is denoted by $M\leq\min(M_a,M_p)$.

In the present formulation, the active party exclusively holds the labels $Y$ and produces the final prediction, while the passive party holds features only. The two roles are therefore:
\begin{itemize}
    \item \textbf{Active party.} Holds $X_a$, the labels $Y$, and the hybrid quantum-classical model $\mathcal{Q}_a$ parameterized by $\theta_a$.
    \item \textbf{Passive party.} Holds $X_p$ only and trains a local model $\mathrm{Mod}_p$ under SBVFL. In the experiments, $\mathrm{Mod}_p$ is an ensemble of classical random forest models, one per synthetic label, with $Q$ the privacy multiplier.
\end{itemize}
 
The active party's objective is to minimize the empirical risk over the intersected sample set. The prediction for sample $i$ combines the output of the active-party model with the contribution returned by the passive party,
\begin{equation}
\hat{y}^{(i)} = g\bigl(\mathcal{Q}_a(x_a^{(i)}; \theta_a), \; o_p^{(i)}; \, W\bigr),
\qquad
\mathcal{L} = \frac{1}{M} \sum_{i=1}^{M} \ell\bigl(\hat{y}^{(i)}, \, y^{(i)}\bigr),
\label{eq:vfl-objective}
\end{equation}
where $\ell$ is the classification loss, $x_a^{(i)}$ is the active-party feature vector, $y^{(i)}$ is the corresponding true label, $o_p^{(i)}$ is the contribution returned by the passive party, and $g$ is the aggregation applied at the server, with parameters $W$. Under SBVFL, $\mathcal{Q}_a$ takes $x_a^{(i)}$ alone as its input, and $\theta_a$ and $W$ are fitted in separate steps, as set out in Section~\ref{sec:bvfl}.
 
\section{Privacy-Preserving ML Solution}
\label{sec:ml-solution}
 
This section defines the centralized reference and the federated protocols used in the experiments, including standard VFL and SBVFL.
 
\subsection{Classical ML Approach}
\label{sec:classical-ml}
 
For the non-private centralized reference, the two feature blocks are pooled and a single classical model $f_\theta$ is trained on $X=[X_a,X_p]$:

\begin{equation}
\min_{\theta} \; \mathcal{J}(\theta) = \frac{1}{M} \sum_{i=1}^{M} \ell\bigl(f_\theta(x_a^{(i)}, x_p^{(i)}), \, y^{(i)}\bigr).
\label{eq:centralized}
\end{equation}

This reference removes the privacy constraints imposed by VFL and is used only to quantify accuracy recovered through collaboration.

\begin{figure}[h]
  \centering
  \includegraphics[width=0.6\linewidth]{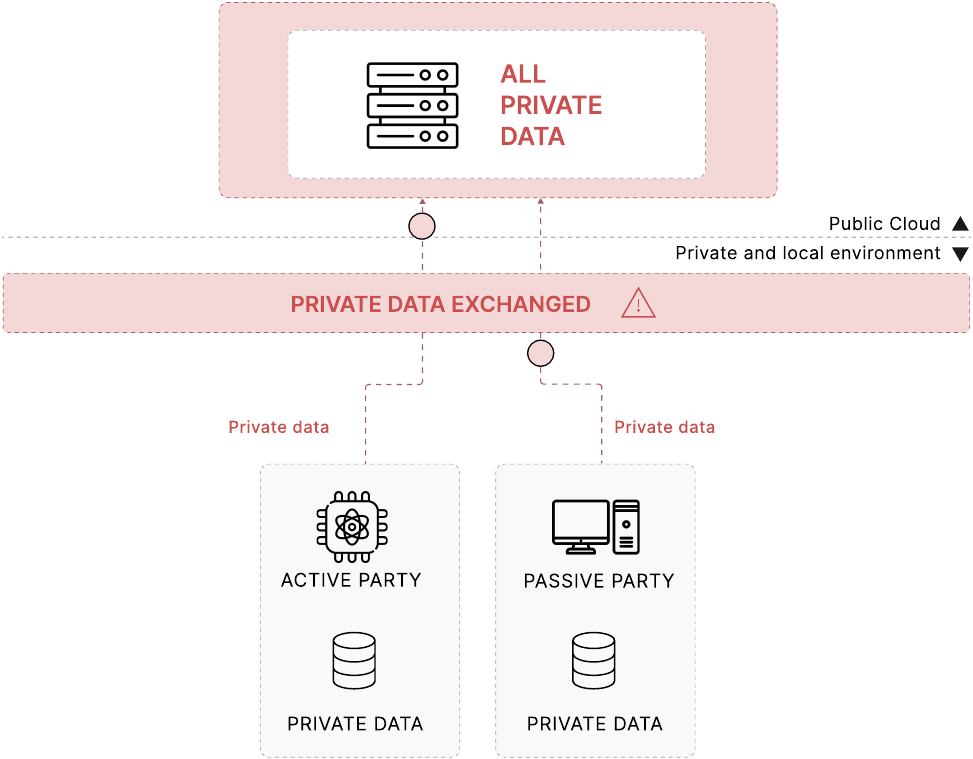}
  \caption{Architecture of the Centralized baseline.}
  \label{fig:centralized}
\end{figure}

\subsection{FL protocols}
\label{sec:bvfl}

FL allows multiple parties to train a shared predictive system while retaining their raw data locally~\cite{mcmahan2017communication}. Depending on the data partition, a protocol may exchange model updates, intermediate representations, or other training quantities rather than raw features. In horizontal FL (HFL), parties share the same feature space but possess different samples; in VFL, they possess complementary feature subsets for overlapping samples. We focus on VFL and, specifically, on the SBVFL variant.

In standard VFL, the passive party computes a local representation
\begin{equation}
h_p^{(i)} = f_p\bigl(x_p^{(i)}; \phi\bigr),
\end{equation}
and transmits it to the active party or a coordinating server. The active party combines this representation with its own representation $h_a^{(i)}=f_a(x_a^{(i)};\theta_a)$, computes the loss, and can return the gradient with respect to $h_p^{(i)}$ so that the passive model can be updated. The parameters of the two local models are not averaged or otherwise merged; each party updates its own parameters. Standard VFL therefore avoids raw-feature exchange but requires repeated communication of label-dependent training quantities. 

\begin{figure}[h]
  \centering
  \includegraphics[width=0.5\linewidth]{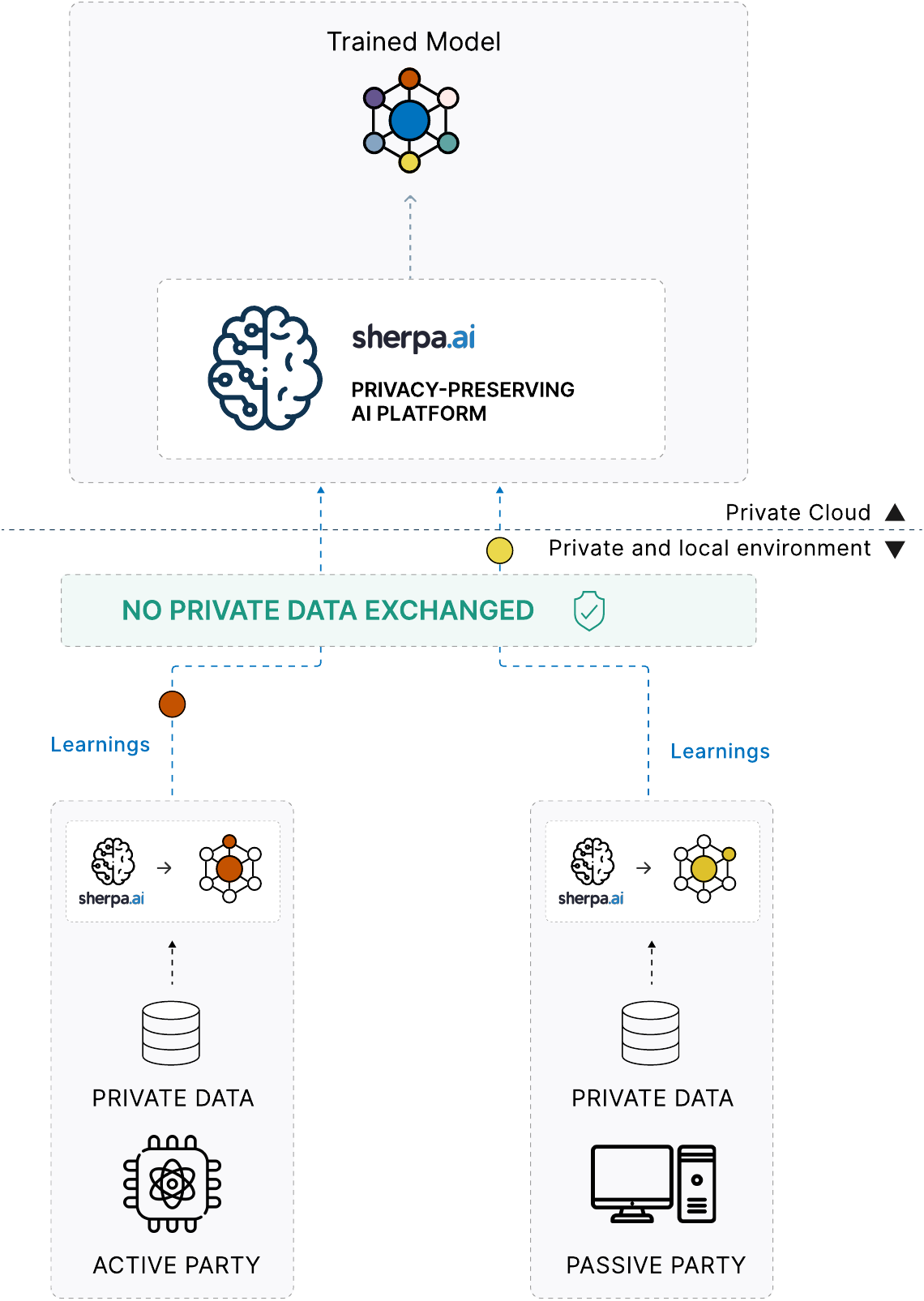}
  \caption{Scheme for the VFL experiment with the FL Platform.}
  \label{fig:bvfl-scheme}
\end{figure}

SBVFL~\cite{acero2025sbvfl} removes this recurring gradient channel: instead of receiving label-dependent gradients, the passive party trains on synthetic targets that the active party generates from the true labels and sends once per training sample. This section states only the steps needed to follow the experiments; the paradigm itself, its communication accounting, and its privacy analysis are given in Acero et al.~\cite{acero2025sbvfl}.

A server, logically embedded within the active party, associates $Q$ synthetic labels with each class and retains the inverse mapping. For each sample it draws one synthetic label of that sample's class, and the resulting synthetic target $z^{(i)}$ is transmitted to the passive party once, before local training. Following~\textcite{acero2025sbvfl}, $Q$ is called the privacy multiplier: the more synthetic labels share a class, the harder the recovery of the real labels from the transmitted targets becomes.

The passive party trains $\mathrm{Mod}_p$ on its own features against the synthetic targets, without access to $Y$ or to the inverse mapping, and returns its trained outputs $o_p^{(i)}$ once. The server assembles the two parties' outputs into the input of the aggregation $g$ of Equation~\eqref{eq:vfl-objective} and trains $g$ against the true labels, which is where the return to the label space takes place. The active-party model consumes $x_a^{(i)}$ alone, and the passive party's outputs enter only at the aggregation. In the two-party setting used here, the passive party is therefore contacted twice, once to receive the synthetic targets and once to return its outputs.

In the experiments, $\mathrm{Mod}_p$ is an ensemble of independent classical random forest models, one per synthetic label, each fitted to the corresponding coordinate of the synthetic target. Increasing $Q$ increases the number of synthetic labels associated with each class; its effect on downstream accuracy is evaluated empirically in Section~\ref{sec:ablation-privacy}. The two parties may therefore use different model families and parameter spaces. Cross-party weight averaging is not required, irrespective of whether a local model is classical, quantum, or hybrid.

\section{The SMPP Benchmark and Its Vertical Partition}
\label{sec:benchmark}

This section motivates and defines SMPP and analyzes the vertical partition required for blind training.

\subsection{Multiplicative Periodic Structure}
\label{sec:why-multiplicative}

Angle encoding induces a structured function class. A qubit prepared from $\lvert0\rangle$ by $R_Y(\omega x)$ has Bloch components $(\sin\omega x,0,\cos\omega x)$, so a Pauli-$Z$ expectation is sinusoidal in the encoded feature. For a product state, a tensor-product observable factorizes into the product of the corresponding single-qubit expectations:

\begin{equation}
\bigl\langle Z_{q_1} \cdots Z_{q_k} \bigr\rangle
  = \prod_{j=1}^{k} c_{q_j} \cos\bigl(\omega_{q_j} x_{q_j} + \varphi_{q_j}\bigr),
  \qquad \lvert c_{q_j} \rvert \le 1 .
\label{eq:product-readout}
\end{equation}

Constructing a target function whose structure is deliberately matched to a model class's inductive bias, to isolate a measurable advantage for that class, is an established practice in QML: Heimman et al.\cite{heimann2025learning} construct Fourier-structured regression targets to compare circuit ans\"atze, while Huang et al.~\cite{huang2021power} and Liu et al.~\cite{liu2021rigorous} construct a discrete-logarithm-based learning problem to establish a provable quantum speed-up over any classical learner. SMPP follows the same principle, applied to the tensor-product readout of an angle-encoded circuit in a vertically partitioned setting.

A readout of this form represents a product of periodic factors with a parameter count that grows linearly with the number of encoded features. A piecewise-linear classical network instead represents the corresponding alternating decision regions through its linear regions. SMPP exposes this difference in inductive bias.

The construction isolates a target family on which the circuit's inductive bias matches the target. All circuits used here are small enough to be simulated exactly by a classical processor, and the primary comparison is the trainable parameter count required to attain a specified accuracy.

\subsection{Task Definition}
\label{sec:task-definition}

Let $d = 4$ and $x \sim \mathrm{Uniform}\bigl([-\pi, \pi]^4\bigr)$. Define the two
\emph{half-signals}
\begin{equation}
A(x) = \sin(\omega x_1)\,\sin(\omega x_2), \qquad
B(x) = \sin(\omega x_3)\,\sin(\omega x_4),
\label{eq:half-signals}
\end{equation}
and the label
\begin{equation}
y = \ind\bigl[A(x) + B(x) > 0\bigr],
\label{eq:label}
\end{equation}

\begin{figure}[h]
  \centering
  \includegraphics[width=\linewidth]{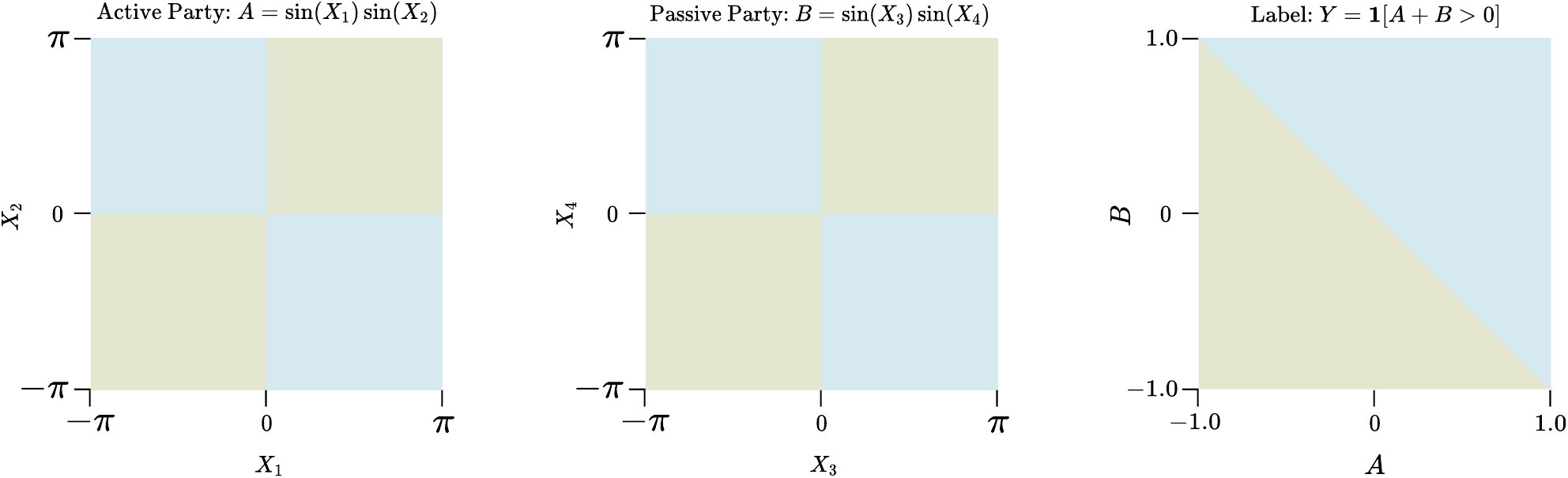}
  \caption{The SMPP task: the two half-signals and the label.}
  \label{fig:task}
\end{figure}

A fraction $\eta$ of labels is then flipped independently, which caps the attainable test accuracy at $1-\eta$. We use $\omega=1$ and $\eta=0.05$, so the cap is $0.95$. We refer to the resulting benchmark as SMPP: each half is a two-feature multiplicative periodic signal, while the label is determined by the additive combination of the two halves. Each half-signal is a sign-alternating checkerboard over its own two features, so neither half is linearly separable, while the label is the sign of their sum. Figure~\ref{fig:task} shows the two half-signals in its left and center panels and the label in its right panel.

The construction has three properties that are central to the federated experiment.

\begin{enumerate}
  \item[(P1)] \emph{Affine models are uninformative.} The periodic sign structure is symmetric under coordinate reflections, giving class-conditional marginals with matching first and second moments, so no affine rule separates the classes at better than chance. The benchmark therefore isolates the role of nonlinear structure.
  \item[(P2)] \emph{Each half is marginally informative.} Because the label combines the two halves additively, conditioning on the sign of either half shifts the label distribution; in the noiseless construction, $\mathbb{P}(y=1\mid A>0)=3/4$. This property is required for passive-party learning from synthetic targets under SBVFL.
  \item[(P3)] \emph{Neither half is sufficient.} A rule based on $A$ alone reaches only $3/4$ accuracy on noiseless labels, whereas access to both halves permits perfect classification. Thus the partition creates a nonzero and quantifiable collaboration gap.
\end{enumerate}

\subsection{Vertical Partition and Additive Label Structure}
\label{sec:partition-design}

The active party receives $(x_1,x_2)$ and the passive party receives $(x_3,x_4)$, so each party owns one complete half-signal. Each local view is therefore itself a two-feature multiplicative periodic task, matched to a two-qubit tensor-product readout.

The additive label rule in Equation~\eqref{eq:label} is essential to the blind protocol. Under SBVFL, the passive party trains only from its own features and the synthetic targets, so its feature block must retain marginal information about the true label. By contrast, for the multiplicative alternative $y=\ind[AB>0]$,
\begin{equation}
\mathbb{P}(y=1\mid A>0)=\mathbb{P}(y=1\mid A<0)=\tfrac{1}{2},
\end{equation}
so each party's feature block is marginally independent of the label. The passive model could then learn no label information from its synthetic targets. SMPP therefore places the multiplicative structure within each party's feature block and combines the two blocks additively. This partition design is a general consideration when constructing blind vertical benchmarks.

\section{Models}
\label{sec:models}

All models use the same preprocessing and are trained under the common procedure described in Section~\ref{sec:experiments}.

\subsection{Hybrid Quantum-Classical Model}
\label{sec:hybrid-model}

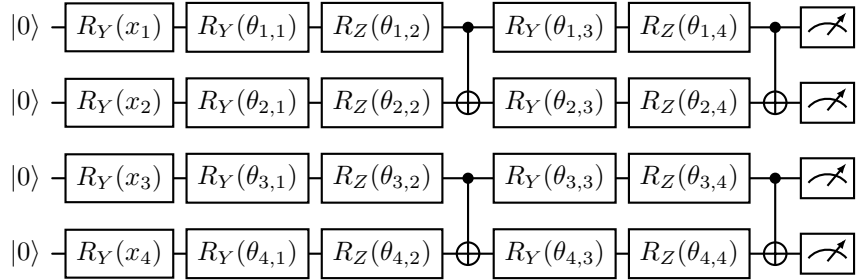
\begin{figure}[h]
  \centering
  \begin{quantikz}[column sep=5pt, row sep=10pt]
    \lstick{$\lvert 0\rangle$} & \gate{R_Y(x_1)} & \gate{R_Y(\theta_{1,1})} & \gate{R_Z(\theta_{1,2})} & \ctrl{1} & \gate{R_Y(\theta_{1,3})} & \gate{R_Z(\theta_{1,4})} & \ctrl{1} & \meter{} \\
    \lstick{$\lvert 0\rangle$} & \gate{R_Y(x_2)} & \gate{R_Y(\theta_{2,1})} & \gate{R_Z(\theta_{2,2})} & \targ{}  & \gate{R_Y(\theta_{2,3})} & \gate{R_Z(\theta_{2,4})} & \targ{}  & \meter{} \\
    \lstick{$\lvert 0\rangle$} & \gate{R_Y(x_3)} & \gate{R_Y(\theta_{3,1})} & \gate{R_Z(\theta_{3,2})} & \ctrl{1} & \gate{R_Y(\theta_{3,3})} & \gate{R_Z(\theta_{3,4})} & \ctrl{1} & \meter{} \\
    \lstick{$\lvert 0\rangle$} & \gate{R_Y(x_4)} & \gate{R_Y(\theta_{4,1})} & \gate{R_Z(\theta_{4,2})} & \targ{}  & \gate{R_Y(\theta_{4,3})} & \gate{R_Z(\theta_{4,4})} & \targ{}  & \meter{}
  \end{quantikz}
  \caption{The quantum layer of the centralized model.}
  \label{fig:circuit}
\end{figure}

The active party uses a variational quantum circuit followed by a trainable classical linear head. The circuit returns a small vector of Pauli-$Z$ correlators, which the classical head maps to a class probability. This is the standard pattern of a quantum layer embedded within a classical network; the specific ansatz is shown in Figure~\ref{fig:circuit}. Each qubit encodes one feature through $R_Y(x_q)$ and carries two layers of trainable $R_Y R_Z$ rotations, and a CNOT couples the two qubits of each pair, so the entangling pattern stays inside a party's feature block (Section~\ref{sec:partition-design}). The readout is the pair of correlators $\langle Z_1 Z_2 \rangle$ and $\langle Z_3 Z_4 \rangle$.

Formally, with $n = d$ qubits, $L$ layers and observables
$\mathcal{O} = \{Z_1 Z_2, \, Z_3 Z_4\}$, the model computes
\begin{equation}
\hat{y}(x) = \sigma\Bigl( w^{\top} \bigl[ \langle Z_1 Z_2 \rangle_{\psi(x;\theta)}, \;
                                          \langle Z_3 Z_4 \rangle_{\psi(x;\theta)} \bigr]
                          + b \Bigr),
\end{equation}
where $\psi(x; \theta)$ is the state prepared by the circuit of Figure~\ref{fig:circuit}, $\sigma$ is the logistic function, and $(w,b)$ are the head parameters. With one qubit per input feature, the circuit consumes the features directly, so the parameter count comprises the circuit rotations and the classical head.

The count is therefore fixed by the number of encoded features and by the output
convention of the training loop. With $n$ encoded features the circuit carries $4n$
angles, two rotations per qubit in each of the $L=2$ layers, and the head carries one
weight per readout correlator for each output, plus one bias per output. The
centralized model encodes four features and reads two correlators, giving $16+3=19$
trainable parameters. A single party's two-feature view uses a two-qubit circuit with
the single correlator $\langle Z_1Z_2\rangle$, giving $8+2=10$. The federated active
party uses that same two-qubit circuit, but the federated training loop scores classes
with a logit vector, so its head maps the correlator to two logits and the count is
$8+4=12$. The two additional parameters express the same decision function
redundantly: the decision depends on the difference of the two logits, which is again
an affine function of the correlator.

Expectation values are evaluated by exact state-vector simulation. For four qubits, the state contains 16 amplitudes, and gradients are obtained by automatic differentiation through the state vector. 

\subsection{Classical Baselines}
\label{sec:baselines}

We use two classical baselines. A fully connected network with one hidden ReLU layer is evaluated at the hidden widths $h\in\{10,50,200\}$ in every scenario. The trainable-parameter count of a given width depends on the input dimension and on the output convention: 41, 201 and 801 on a party's two-feature view, 61, 301 and 1201 on the pooled four features, and 52, 252 and 1002 for the federated active party, whose head emits two logits. The sweep therefore provides a capacity range rather than a single classical reference.

An ensemble of 300 random forest trees is included as a nonlinear tree-based baseline. Its size is reported as the total number of tree nodes rather than as a trainable parameter count and is therefore kept separate from the neural-network and circuit parameter counts.

\section{Experimental Protocol}
\label{sec:experiments}

We now describe the experimental protocol underlying the results of Section~\ref{sec:results}, comprising the three training scenarios evaluated for each model family and the training, preprocessing, and evaluation procedure common to all of them.

\subsection{Training Scenarios}
\label{sec:scenarios}

The protocol evaluates three scenarios, listed here in the order in which
Section~\ref{sec:results-scenarios} reports them. The underlying data, splits, optimization
procedure, and hyperparameter selection are held fixed; the scenarios differ in the feature
columns available to the model and in whether the two parties collaborate.

\begin{enumerate}
  \item \textbf{Local active-party.} The active party trains using $(x_1,x_2)$ only, providing the non-collaborative reference for the active side.
  \item \textbf{SBVFL.} The labels are encoded at multiplier $Q$, the passive party fits its random forest ensemble on $(x_3,x_4)$ and returns its outputs once, the active-party model consumes $(x_1,x_2)$ alone, and the server's aggregation combines the two parties' outputs into the prediction.
  \item \textbf{Centralized.} The four features $(x_1,x_2,x_3,x_4)$ are pooled and a single model is trained on the combined data. This is the non-private reference.
\end{enumerate}

For a local party view, the hybrid component is reduced to a two-qubit circuit with one readout correlator $\langle Z_1Z_2\rangle$ and 10 trainable parameters, and in the federated scenario that same circuit carries the two-logit head required by the federated training loop, giving 12; the classical baselines are reduced to the corresponding input dimension in both cases. Otherwise, the model specification is unchanged. Within a scenario, all models receive the same feature columns and are therefore compared on identical inputs; across scenarios the input dimension differs, so the centralized experiment selects the model family used by the active party rather than a fixed parameter count.

Federated experiments are executed on \sherpa's platform through its blind vertical training interface. The platform supplies the PSI step, label encoding, passive-party fitting, and blind aggregation; this work specifies the dataset and the active-party model. The intersection is partitioned so that the federated training fold contains the same 2000 rows used by the centralized and local experiments, enabling direct comparison of test accuracies.

\subsection{Protocol}
\label{sec:protocol}

One SMPP instance with 2000 training rows and 20,000 test rows is drawn once and reused by every experiment, so the centralized comparison, the local baseline, and the federated runs are evaluated on the same rows, with each party's local view given by the corresponding column block of the pooled view. Each configuration is then evaluated over 10 training seeds, which determine the model initialization and batch order only. The large test set keeps test-sampling uncertainty below the seed-to-seed variation of interest.

All model families use the same rows, split, and preprocessing. We report the mean test accuracy across the 10 training seeds with its standard deviation, and compare models by the difference in their means. The standard deviation measures variation over training seeds at a fixed benchmark instance; it does not cover variation across independently drawn benchmark instances.

All gradient-trained models use binary cross-entropy loss, AdamW, batch size 128, a 300-epoch cap, and the same early-stopping rule. Training monitors a stratified 20\% inner-validation holdout, uses a patience of 30 epochs, and restores the best validation state.

Learning rates are selected separately for each model family from $\{0.3,0.1,0.03,0.01,0.003\}$ using inner validation and tuning seeds that are disjoint from the reporting seeds. The selected rates are then fixed for all reported experiments.

All models use the same preprocessing rule: each feature is mapped to $[-\pi,\pi]$ using the minimum and maximum of the training fold. This scaling is required by the angle encoding and is applied identically to the classical baselines.

Test accuracy is the primary metric, with AUC reported as a secondary metric. Parameter counts are obtained programmatically from the model definitions. In the federated scenario they cover the active party's own model; the server's aggregation is supplied by the platform and is identical for every active-party model compared, since all of them emit two class scores.

The simulations use PyTorch $2.12$ with float64/complex128 arithmetic, NumPy $2.4.6$, scikit-learn $1.9.0$, and Python $3.12$. The quantum circuits use Qiskit $1.4.4$. Experiments are executed on a single CPU core per configuration.

\section{Results}
\label{sec:results}

In this section we report the effect of collaboration on classification accuracy across the three training scenarios, followed by an ablation of the passive party's privacy multiplier.

\subsection{The Three Training Scenarios}
\label{sec:results-scenarios}

\begin{figure}[h]
  \centering
  \includegraphics[width=0.58\linewidth]{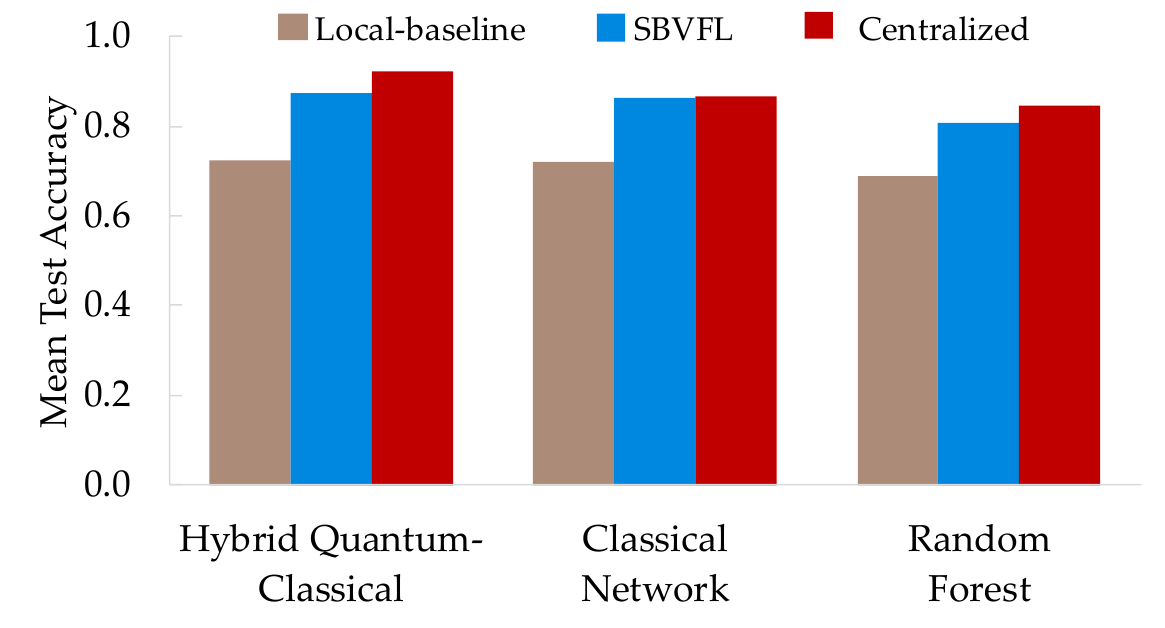}
  \caption{Mean test accuracy of the three model families in the three scenarios.}
  \label{fig:scenarios}
\end{figure}

Table~\ref{tab:scenarios} and Figure~\ref{fig:scenarios} report the three scenarios of Section~\ref{sec:scenarios} for each model family. Every entry is the mean test accuracy over the 10 training seeds with its standard deviation in parentheses, measured on the single
benchmark instance of Section~\ref{sec:protocol}. The classical network is the best width of the sweep in each scenario. In the local and federated scenarios the hybrid model is the two-qubit circuit with one readout correlator; in the centralized scenario, it is the four-qubit model of Figure~\ref{fig:circuit}. The Parameters row reports the number of trainable parameters for the hybrid and classical-network families, obtained programmatically from the model definitions (Section~\ref{sec:experiments}), and the total number of tree nodes for the random forest (average over the ten seeds).


\begin{table}[h]
  \centering
  \small
    \resizebox{\textwidth}{!}{%
  \begin{tabular}{l l c c c}
    \toprule
    \textbf{Scenario} & \textbf{Metric} & \textbf{Hybrid quantum-classical} & \textbf{Classical network} & \textbf{Random forest} \\
    \midrule
    \multirow{2}{*}{Local baseline, active party ($x_1,x_2$)}
      & Accuracy   & \textbf{0.7227} (0.0021) & 0.7223 (0.0016) & 0.6895 (0.0011) \\
      & Parameters & 10               & 41               & 207{,}736 nodes \\
    \midrule
    \multirow{2}{*}{SBVFL}
      & Accuracy   & \textbf{0.8757} (0.0001) & 0.8623 (0.0017) & 0.8065 (0.0009) \\
      & Parameters & 12               & 1002             & 673,276 nodes \\
    \midrule
    \multirow{2}{*}{Centralized, both parties ($x_1,\dots,x_4$)}
      & Accuracy   & \textbf{0.9216} (0.0056) & 0.8663 (0.0068) & 0.8462 (0.0011) \\
      & Parameters & 19               & 1201             & 140{,}741 nodes \\
    \bottomrule
  \end{tabular}}
  \caption{Test accuracy (mean, with standard deviation in parentheses) and trainable-parameter count in the three training scenarios. The random forest's size is reported as its total number of tree nodes rather than a trainable-parameter count, since it has no trainable parameters in the usual sense.}
  \label{tab:scenarios}
\end{table}

Under the active party's feature view, the hybrid model reaches $0.7227$ with 10 trainable parameters, the best classical network reaches $0.7223$ with 41, and the random forest reaches $0.6895$ with 207,736 tree nodes. The two parametric families agree to within $0.001$, although their trainable-parameter counts differ by a factor of four. Widening the network does not improve its local accuracy: it reaches $0.7223$, $0.7187$, and $0.7183$ at hidden widths $10$, $50$, and $200$, respectively, corresponding to 41, 201, and 801 trainable parameters. The local accuracy is therefore limited primarily by the information contained in the feature block, as property~(P3) of Section~\ref{sec:benchmark} anticipates. The partition is symmetric by construction, with Equation~\eqref{eq:half-signals} assigning each party a half-signal of the same form, so neither party holds a privileged feature block.

With the four features pooled, the hybrid model reaches $0.9216$ with 19 trainable parameters, within $0.028$ of the $0.95$ cap imposed by the label noise. The best classical network reaches $0.8663$ with 1201 parameters, and the 300-tree forest reaches $0.8462$ with 140,741 nodes. The margin over the network, $+0.0554$, is about eight times the larger of the two standard deviations, $0.0056$ and $0.0068$, while the margin over the forest is $+0.0755$. This comparison motivates the selection of the hybrid model as the active party in the federated experiment. 

The hybrid model improves from 0.7227 in the local scenario to 0.9216 in the centralized one. SBVFL reaches 0.8757 with 12 trainable parameters in the active-party network, comprising eight circuit rotations and a four-parameter head (Section~\ref{sec:hybrid-model}), substantially improving on the local baseline and closely approaching the centralized reference. The hybrid model is also the most accurate active party tested, with margins of $+0.0134$ over the best classical network at 1002 parameters ($h=200$), $+0.0191$ over $h=50$, $+0.0496$ over $h=10$, and $+0.0692$ over the forest (0.8065, 673{,}276 nodes). Each margin exceeds the seed-to-seed standard deviation of the corresponding comparison. The SBVFL forest is markedly larger than the forests trained on the same two-feature view in the local scenario (207,736 nodes) or on the pooled four-feature view in the centralized scenario (140,741 nodes), which we attribute to the platform's training loop rather than to the two-feature input itself, as discussed below. The ordering of the model families is the same as in the centralized scenario. The passive party observes neither the true labels nor label-dependent gradients and is contacted twice during the protocol.

The local baseline and the federated configuration differ in more than the passive party's contribution: the passive features enter through $o_p^{(i)}$, the server fits its aggregation against the true labels, and the active-party model is trained through the platform's loop rather than through the local harness. The experiments vary these factors together and do not separate them, so interpreting the gain as the value of the passive party's information is a property of the experimental design rather than a measured decomposition. An aggregation applied to the active party's output alone is a function of that output, so any gain from that source is bounded by the accuracy attainable from the active model's own score.

\subsection{Ablation: The Passive Party's Privacy Multiplier}
\label{sec:ablation-privacy}

\begin{figure}[h]
  \centering
  \includegraphics[width=0.58\linewidth]{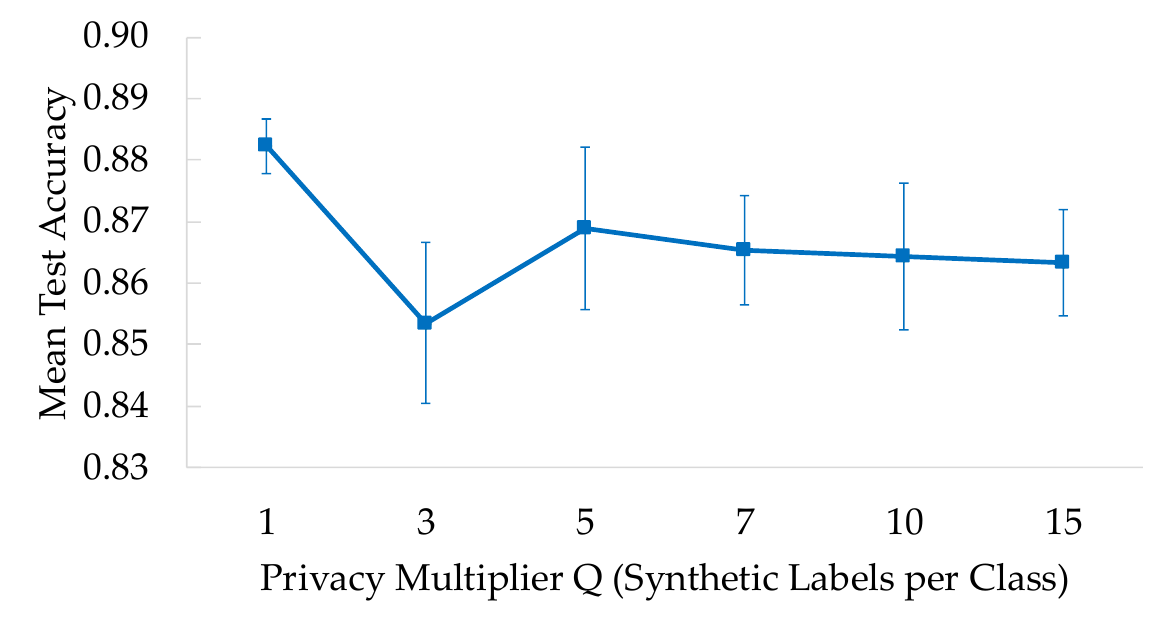}
  \caption{Mean test accuracy as a function of the privacy multiplier $Q$, with error bars denoting standard deviation across seeds.}
  \label{fig:privacy}
\end{figure}

The privacy multiplier $Q$ controls the number of synthetic labels associated with each class. Figure~\ref{fig:privacy} varies $Q$ while keeping the remaining protocol fixed; every setting is measured on the same benchmark instance and the same ten training seeds, and every entry is the mean test accuracy with its standard deviation. The quantity varied is the platform's privacy-multiplier setting. No privacy guarantee and no privacy loss is quantified in this work, so these measurements describe accuracy against $Q$ and not a privacy-utility curve. In this sweep the synthetic-label draw follows the training seed, so each mean averages over ten label-encoding realizations, whereas the federated scenario of Section~\ref{sec:results-scenarios} fixes one realization together with the benchmark instance. The two settings therefore measure different quantities at the same multiplier, $0.8653$ here against $0.8757$ in Table~\ref{tab:scenarios}, and the sweep is to be read within itself.


At $Q=1$, the active-party accuracy exceeds that at the reference setting $Q=7$ by $0.0170$, against standard deviations of $0.0044$ and $0.0089$. For $Q=5,10,15$, the differences from $Q=7$ are $+0.0035$, $-0.0011$, and $-0.0019$, respectively, each smaller than the corresponding seed-to-seed standard deviations. The $Q=3$ setting is lower by $0.0118$, so accuracy is not monotone in $Q$ over the tested values.

For the tested settings above the weakest multiplier, accuracy is approximately flat in $Q$. The strongest multiplier tested is at least as accurate as the weakest: $Q=15$ exceeds $Q=3$ by $0.0099$ in the mean. Among the tested settings, only $Q=1$ differs from the default by an amount larger than the corresponding seed-to-seed variation.

\section{Discussion}
\label{sec:discussion}

The main result of the manuscript is the accuracy recovered under blind federated training. SBVFL substantially improves on active-party local training and closely approaches the non-private centralized reference, while preserving the ordering of the active-party model classes observed in the centralized experiment. The privacy ablation further indicates that, over the tested settings, increasing the multiplier from $Q=3$ to $Q=15$ does not reduce accuracy; the mean increases by $0.0099$ across the ten seeds.

The partition analysis shows that local accuracy is determined by the information available in each feature block. On the active party's view the model families plateau together, at $0.7227$ and $0.7223$ for the circuit and the classical network, and added capacity does not move them, while the pooled hybrid model reaches $0.9216$, yielding a collaboration gap of $0.1989$. This interpretation depends on using a task for which the local view is itself close to saturation.

The partition analysis also yields a general benchmark-design principle. The multiplicative structure exploited by the tensor-product readout must be contained within each party's feature block rather than introduced only through an interaction across parties. A multiplicative combination across the partition would make each local feature view marginally independent of the label and would therefore be incompatible with the passive-party learning step of SBVFL. SMPP separates these roles by using multiplicative structure within each block and additive composition across blocks.

The active-party model choice follows from this structural match rather than from a general advantage of quantum models. When the target contains products of periodic functions of distinct features, the tensor-product readout represents that form directly with a compact parameterization, whereas a piecewise-linear network approximates the same structure through a collection of linear regions. On SMPP, this produces a $0.055$ accuracy margin over the 1201-parameter classical network and a $0.075$ margin over the 300-tree forest, both at least eight times the larger of the two standard deviations in the corresponding comparison.

Parameter efficiency is relevant operationally because gradient-based training on hardware requires repeated circuit evaluations for the trainable parameters. Reducing the model from 1201 to 19 trainable parameters therefore reduces the number of parameter-dependent evaluations required for optimization, although this study does not measure hardware wall-clock performance. The accounting is the parameter-shift one, in which each trainable circuit parameter contributes evaluations at every optimization step while the classical head is differentiated analytically. The comparison is between trainable-parameter counts; the random forest has none and is sized by its tree nodes, so it enters as an accuracy reference rather than as a point on the parameter axis. The reported circuit uses four qubits and a shallow architecture, but all numerical results here are obtained with exact classical simulation.

These results have some limitations. First, SMPP's measured parameter efficiency reflects targets whose structure matches this multiplicative periodic form; a second benchmark instance kept the ordering of model families unchanged, with accuracy shifting by at most $0.0123$. Second, the comparison concerns a four-qubit circuit that is exactly
classically simulable, and the federated experiments run all parties within a single process without network transport, so the reported results reflect predictive behavior rather than distributed runtime. Third, validating these findings on additional, real-world datasets remains necessary to establish the broader effectiveness of this
approach. Finally, the results assume complete sample overlap and consider only $\omega=1$ with a single label-noise level; larger circuits can also suffer from gradient concentration~\cite{mcclean2018barren}, and how parameter efficiency behaves at larger register sizes or depths remains to be established~\cite{abbas2021power}.

\section{Conclusion}
\label{sec:conclusions}

We have evaluated a hybrid quantum-classical model as the active party in a two-party SBVFL protocol on the SMPP benchmark. Our experiments show that SBVFL improves accuracy from 0.7227 (local baseline) to 0.8757, closely approaching non-private centralized training, with the hybrid quantum-classical model reaching this accuracy using only 12 trainable parameters versus 1002 (classical network) and 673,276 nodes (random forest). A privacy-multiplier ablation further shows that accuracy does not systematically degrade as $Q$ (the number of synthetic labels per class) increases.

These findings confirm that FL enables privacy-preserving collaboration between a hybrid quantum-classical active party and a classical passive party without centralizing raw data, while the quantum model retains its accuracy advantage under this protocol. This makes parameter-efficient quantum models combined with blind vertical federation a promising direction for multi-party ML in settings such as chemistry simulation, financial modeling, and image classification tasks, where modern quantum hardware and strict data-sharing constraints coexist.

\section*{Contributions and Acknowledgments} \label{app:A}

Carlos Cano 

Daniel M. Jimenez-Gutierrez

Diego Sal


Georgios Kellaris

Joaquin del Rio

Oleksii Sliusarenko

Xabi Uribe-Etxebarria

\vspace{5mm}
The authors are presented in alphabetical order by first name.

\printbibliography

\end{document}